\documentclass[conference, a4paper]{IEEEtran}
\IEEEoverridecommandlockouts
\usepackage{cite}
\usepackage{amsmath,amssymb,amsfonts}
\usepackage{algorithmic}
\usepackage{graphicx}
\usepackage{textcomp}
\usepackage{xcolor}
\def\BibTeX{{\rm B\kern-.05em{\sc i\kern-.025em b}\kern-.08em
    T\kern-.1667em\lower.7ex\hbox{E}\kern-.125emX}}

\usepackage{hyperref}       %
\usepackage{url}            %
\usepackage{booktabs}       %
\usepackage{nicefrac}       %

\hypersetup{
    colorlinks, linkcolor={dark-red},
    citecolor={dark-blue}, urlcolor={medium-blue}
}

\newcommand\copyrighttext{%
© © 2024 IEEE. Personal use of this material is permitted. Permission from IEEE must be obtained for all other uses, in any current or future media, including reprinting/republishing this material for advertising or promotional purposes, creating new collective works, for resale or redistribution to servers or lists, or reuse of any copyrighted component of this work in other works.}
\newcommand\copyrightnotice{%
	\begin{tikzpicture}[remember picture,overlay]
		\node[anchor=south,yshift=10pt] at (current page.south) {\fbox{\parbox{\dimexpr\textwidth-\fboxsep-\fboxrule\relax}{\copyrighttext}}};
	\end{tikzpicture}%
}

\usepackage{subcaption}
\usepackage{multicol, blindtext, pgfplots, tikz, mathtools}

\definecolor{dark-green}{rgb}{0,0.5,0}
\definecolor{light-green}{HTML}{D6F6DD}
\definecolor{lavender-pink}{HTML}{DAC4F7}
\definecolor{salmon-pink}{HTML}{F4989C}
\definecolor{creme-creme}{HTML}{EBD2B4}
\definecolor{baby-blue}{HTML}{12A1BA}
\definecolor{dark-red}{rgb}{0.4,0.15,0.15}
\definecolor{dark-blue}{rgb}{0.15,0.15,0.8}
\definecolor{medium-blue}{rgb}{0,0,0.5}

\pgfplotsset{every axis/.append style={
                    ylabel near ticks,
                    xlabel near ticks,
                    tick pos=left
                    }}

\pgfplotscreateplotcyclelist{color-list}{
salmon-pink,every mark/.append style={fill=salmon-pink!80!black},mark=*\\
baby-blue,every mark/.append style={fill=baby-blue!80!black},mark=*\\
dark-green,every mark/.append style={fill=dark-green!80!black},mark=*\\
creme-creme,every mark/.append style={fill=creme-creme!80!black},mark=*\\
}

\pgfplotscreateplotcyclelist{color-list2}{
creme-creme,every mark/.append style={fill=creme-creme!80!black},mark=*\\
baby-blue,every mark/.append style={fill=baby-blue!80!black},mark=*\\
}

\begin{document}

\title{On the Road to Clarity: Exploring Explainable AI %
for World Models in a Driver Assistance System\\
\thanks{Partially funded by the Federal Ministry of Education and Research (BMBF), projects `NEUPA', grant 01IS19078, and `KI-IoT', grant 16ME0091K.}
\thanks{ We thank Robert, Ellen, and Anita de Mello Koch for their insights and advice, and for providing their source code. %
}
}

\author{
\IEEEauthorblockN{Mohamed Roshdi\IEEEauthorrefmark{1}, Julian Petzold\IEEEauthorrefmark{1}, Mostafa Wahby\IEEEauthorrefmark{1}, Hussein Ebrahim\IEEEauthorrefmark{1}, Mladen Berekovic\IEEEauthorrefmark{1}, Heiko Hamann\IEEEauthorrefmark{2}}
\IEEEauthorblockA{\IEEEauthorrefmark{1}Institute of Computer Engineering\\ University of L\"ubeck, L\"ubeck, Germany\\ Email: petzold@iti.uni-luebeck.de}
\IEEEauthorblockA{\IEEEauthorrefmark{2}Department of Computer and Information Science\\ University of Konstanz, Konstanz, Germany%
}
}

\maketitle
\copyrightnotice
\begin{abstract}

In Autonomous Driving (AD) transparency and safety are paramount, as mistakes are costly. However, neural networks used in AD systems are generally considered black boxes. As a countermeasure, we have methods of explainable AI (XAI), such as feature relevance estimation and dimensionality reduction. Coarse graining techniques can also help reduce dimensionality and find interpretable global patterns. A~specific coarse graining method is Renormalization Groups from statistical physics. It has previously been applied to Restricted Boltzmann Machines (RBMs) to interpret unsupervised learning. We refine this technique by building a transparent backbone model for convolutional variational autoencoders (VAE) that allows mapping latent values to input features and has performance comparable to trained black box VAEs. 
Moreover, we propose a custom feature map visualization technique to analyze the internal convolutional layers in the VAE to explain internal causes of poor reconstruction that may lead to dangerous traffic scenarios in AD applications. %
In a second key contribution, we propose explanation and evaluation techniques for the internal dynamics and feature relevance of prediction networks. We test a long short-term memory (LSTM) network in the computer vision domain to evaluate the predictability and in future applications potentially safety of prediction models. 
We showcase our methods by analyzing a VAE-LSTM world model that predicts pedestrian perception in an urban traffic situation.

\end{abstract}

\section{Introduction}
\label{section:intro}
Methods based on artificial intelligence (AI) have been shown to have increasing successes when applied to a vast variety of application fields (e.g., healthcare, farming, autonomous vehicles, etc.). 
Many symbolic AI approaches (e.g., rule-based methods) can represent problems in an easily interpretable and human-readable format.
However, machine learning (ML) techniques that have shown more promising performance, such as artificial neural networks (ANNs), follow the subsymbolic paradigm. These techniques are considered `black boxes' that are difficult to interpret and explain.

When designing an ML system, its interpretability and explainability are essential factors because they influence the user's trust and their ability to improve or re-adapt the system. %
A~system is interpretable when represented in a human-understandable format and is considered explainable if humans can comprehend the decisions and predictions of the ML system~\cite{Erasmus2021}.
This led to the emergence of the Explainable Artificial Intelligence (XAI) field, which aims to provide means to help interpreting and explaining ML systems.

In the context of autonomous driving, which is the main focus of this work, XAI is of particular interest due to the continuous increase of automation levels and the necessity to explain, at least retrospectively, the decisions made by large ANNs in dangerous situations.
Here, we present an XAI approach that we showcase in an AD-specific application. We use XAI to explain previously trained ANN models that predict the behavior of vulnerable road users (VRUs; e.g., pedestrians, bicycles)~\cite{yannis2020vulnerable} in urban traffic, as their safety is of highest concern. 
These prediction models can be used to develop an advanced driver assistance system (DAS), for example, as an early warning system that tries to anticipate dangerous situations on second-timescale.
The line of sight between a human driver and pedestrians on the sidewalks can provide information on whether a~pedestrian is planning, for example, to enter the road or perform any dangerous behavior%
~\cite{leveque2020pedestrians}. 
Some prediction models exploit the same visual information of observing close-by VRUs to anticipate dangerous situations.
We collected data from the pedestrian perspective at road crossing scenarios in simulations to train ANN~\cite{petzold2022if}. %
This work is based on a~synthetic environment using the CARLA traffic simulator~\cite{Dosovitskiy17}. 
We trained variational autoencoders (VAE) and long short-term memory (LSTM) networks that can be used to predict the positions and trajectories of VRUs in the near future (e.g., one second).
With today's technology, a system on a vehicle equipped with a~360-degree camera could reconstruct relevant features of the perspective of surrounding VRUs to use them as input for prediction models. The feasibility and efficiency of this approach are increased if multiple cars share their perception (e.g., vehicle-to-X approaches~\cite{li2022v2x}).

\begin{figure*}[ht]
    \centering
    \includegraphics[width=1.0\textwidth]{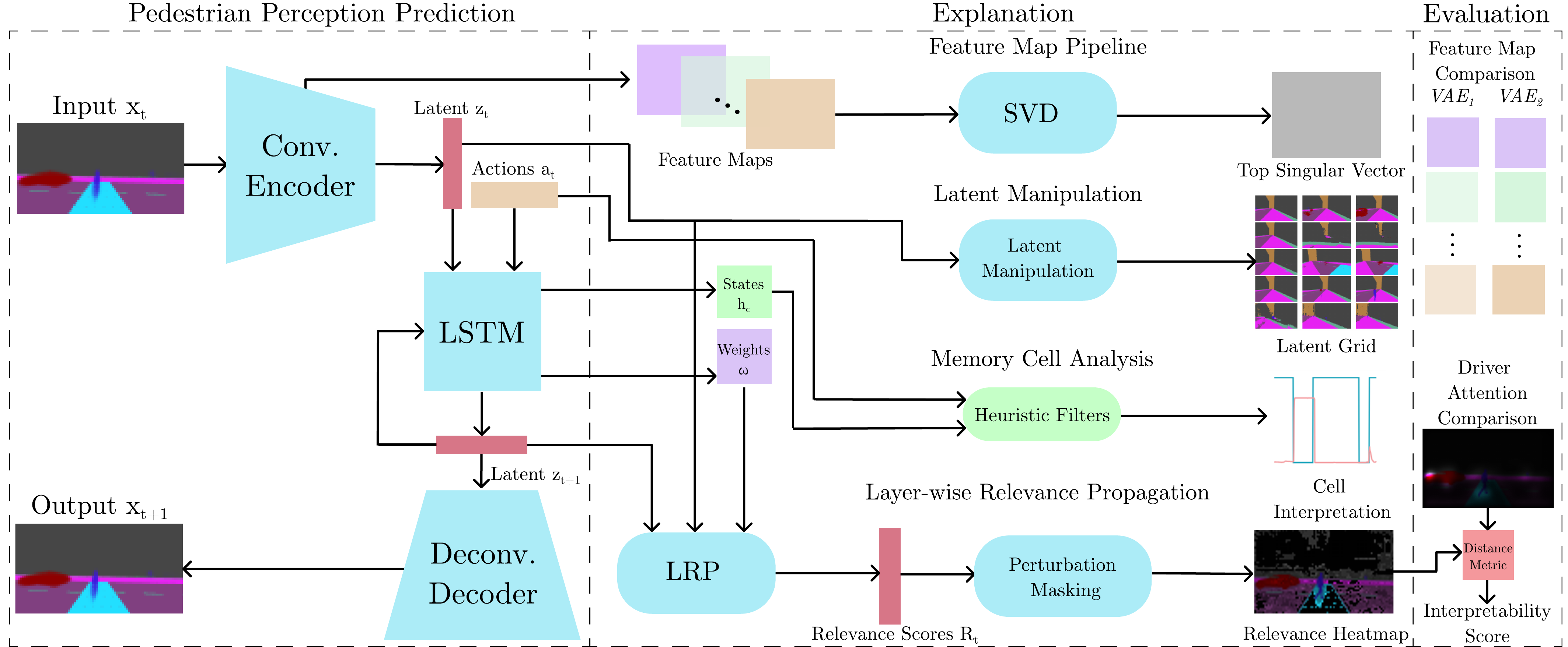}
    \caption{An overview of our XAI system and its components.}
    \label{fig:overview}
\end{figure*}

In this paper, we present four methods and tools to explain both the VAE and LSTM networks of the prediction model from~\cite{petzold2022if} as shown in~Figure~\ref{fig:overview}.
To explain the internal functionality of the convolutional VAE, we develop a 
tool based on activation visualization that visualizes the feature maps and principal components of the convolutional filters.
This provides a visual understanding of the evolution of features through the convolutional layers and filters, enabling the user of the tool to understand the internal functionality of the ConvVAE layers and its generated features. 
Second, we develop a visualization of the learned data manifold. We visualize the range of features encoded by the latent vector in a grid as done by~\cite{vae} and utilized in an interactive tool by \cite{Ha:2018:RWM:3327144.3327171} to explain the mapping between the image features and encoded latent space values.
As the challenge of fully explaining VAEs is vast, we present an alternative approach based on Renormalization Groups (RG)~\cite{koch2} from statistical physics as a third method. It transforms the model to an inherently transparent and interpretable model.
In a fourth approach, we explain LSTMs by correlating the memory cells' hidden states with input features and implementing a custom feature relevance technique to assign relevance scores for input features. 
In an application-focused effort, we test how the LSTM reacts to input frames with domain-specific varying features (e.g., increasing number of pedestrians).  We empirically evaluate the interpretability by comparing LSTM relevance heat maps to human visual attention maps. Our analysis yields a mean Normalized Scanpath Saliency (NSS)~\cite{driverattention} of 0.53, comparing our heat maps to ground truth human visual attention, and reveals abnormal prediction behavior with potential implications for traffic safety. Our explanation and evaluation methods offer a basis for assessing the explainability and predictability of pedestrian perception prediction models that can be applied on different architectures.

\section{Related Work}

To provide explainability for black box models, XAI techniques rely on \emph{interpretations}, a mapping from an abstract domain into a human-understandable domain~\cite{montavon}. %
\cite{xaiconcepts} provide a taxonomy of XAI methodologies based on the insights they provide. Model Simplification techniques aim to compare complex and simplified models to gain insights. 
Deep Network Representation techniques aim to interpret the representation of data in the model. Deep Network Processing techniques provide insights on why certain inputs lead to their observed outputs. 
In this paper, we cover all three of these techniques. %

\paragraph{XAI in Autonomous Driving}
\label{paragraph:Model}
Some advantages of XAI, such as accessibility, confidence, or fairness~\cite{xaiconcepts}, are beneficial for almost all uses of artificial intelligence, but certain application areas, such as %
autonomous driving (AD), are under more scrutiny~\cite{atakishiyev2021towards,omeiza2021explanations,zablocki2022explainability,dong2021image}. %
An error in an AD system %
may cause high costs. %
AD system failure modes may be difficult to comprehend for humans, which reduces trust. For example, in a fatal accident~\cite{ntsb2019}, an autonomous car did not recognize a pedestrian pushing a bicycle. %
To increase safety and to 
alleviate trust issues, %
AD models require insights in causality to build confidence in their mechanisms. Hence, we focus on explaining a model used for AD.
In the domain of AD, vehicle-to-X (V2X)~\cite{li2022v2x} allow cars to communicate with traffic signs, other cars, or even pedestrians. Data obtained this way enables new approaches, such as perceiving traffic participants hidden from the car's line of sight, but visible by the camera of a traffic light. 
V2X approaches might also pose different challenges for XAI, for example, in the domain of machine behavior~\cite{machineBehavior19}. 
A~relevant approach~\cite{petzold2022if} uses action and camera data from the perspective of a pedestrian to predict what the pedestrian will see in future time steps. %

\paragraph{Model-Specific XAI Approaches}
\label{paragraph:related1}
Numerous approaches have focused on developing explainability techniques designed for particular Deep Learning (DL) architectures. Karpathy et al.~\cite{lstmviz} analyzed the functionality of LSTM memory cells in language models, detecting functionalities, such as maintaining the state of long-term dependencies, while Bach et al.~\cite{lrp} proposed Layer-wise Relevance Propagation to assign relevance scores to input features by propagating the model's outputs backwards using redistribution rules. 

To interpret convolutional neural networks (CNNs), \cite{cnnviz} visualized a network's learned features by optimizing random images to maximize the activations of convolutional filters, entire layers, or individual channels. 

Selvaraju et al.~\cite{gradcam}~introduced %
GradCAM, a gradient-based technique that generates a localization map highlighting the relevant areas of an image that influence a classification. Lie et al.~\cite{liu2020towards} extend this gradient-based visualization from classification networks to generative models like the VAE. Muhammad and Yeasin~\cite{eigencam} augmented CAM methods by visualizing the principal components of feature maps.
Cetin et al.~\cite{cetin2023attri} introduce the Attri-VAE, a VAE based on the $\beta$-VAE~\cite{higgins2017beta}. By introducing an attribute regularization term to their loss function, the authors disentangle latent space variables, compelling them to align with predefined image attributes for enhanced interpretability.

\paragraph{DL Interpretation through Renormalization Groups}
\label{sub-subsection:rgIntro}
In statistical physics, Renormalization Groups (RGs) are used to transform complex systems with a high order of parameters into simpler ones with a smaller set of parameters that can describe the general behavior of the system~\cite{scaleinv}. Metha and Schwab~\cite{rgmap} experimentally demonstrated a mapping between RG and Restricted Boltzmann Machines (RBMs). The authors used this mapping to coarse-grain an Ising model, a 2D grid representing the spins of a magnet. Koch et al.~\cite{koch1} explored this connection further by comparing stacked RBMs to a flow of RG operations, where each layer performs a step of RG-like coarse graining. The authors also compared downsampling by convolutional pooling layers to the reduction in dimensionality performed by RG.
Koch et al.~\cite{koch2} present an RG-inspired interpretation of unsupervised learning of %
RBMs and generative models. They compare between the weight matrix of an RBM trained on Ising data %
and an RG using Singular Value Decomposition (SVD)~\cite{svd}. 
Koch et al.~\cite{koch2} noted that the singular vectors with the highest singular values of an SVD of image datasets %
have their support in low frequencies, with little information encoded in high frequencies. This phenomenon is similar to Momentum Space RG, another type of RG that eliminates high momentum modes represented as high frequencies. Koch et al.~\cite{koch2} introduced the RG Machine (RGM) that uses singular vectors of a training set to estimate weights and biases of an RBM. The RGM relies on low frequency modes of singular vectors, produces images similar to those of the RBM, and improves its performance within a few epochs. Hence, the authors interpret the RBM learning process as a process similar to momentum space RG, keeping relevant features represented by low frequency modes of singular vectors.

\section{Methods}
In this section, we present the interpretability methods that we applied to the Convolutional VAE and the LSTM of the pedestrian perception prediction approach in traffic situations~\cite{petzold2022if}.
Our goal is to create an~explainable AI framework that provides human-understandable explanations of the internal processing within the models and the mapping between their inputs and outputs.
\paragraph{Feature Maps Visualization Tool}
\label{paragraph:convVaeMeth}
We developed a tool that visually explains the internal functionality of the ConvVAE model presented in \cite{petzold2022if}. The model encodes pedestrian perception into an abstract 50D latent vector~$z$. This way, the perception is compressed and can be used to predict the next frame with an LSTM model, a type of recurrent neural network for time series prediction~\cite{lstmoriginal}. The input frames are represented in 24 channels corresponding to the 24 semantic classes of a custom CARLA environment. The VAE consists of four convolutional layers that sequentially extract lower-dimensional features from the input image. We introduce a Deep Network Representation \cite{xaiconcepts} tool that interprets the functionality of the convolutional filters and their role in the feature extraction process in two phases. Our feature visualization tool feeds an image to the first convolutional layer and saves the output feature maps at each layer. We apply Min-Max normalization~\cite{Ahsan2021EffectOD} to the feature maps and use them as weighted masks that act on the input image to map the grey-scale features to the RGB features, producing more legible feature maps. Next, the tool uses the feature maps at each layer to compute the SVD \cite{svd} of the images. This allows us to visualize the top singular vectors $v$ of the feature maps representing the principal components and gain a~visual description of the layer's functionality, similar to EigenCAM~\cite{eigencam} but in a regression rather than classification task.
We conducted an experiment to evaluate the effectiveness of our approach in producing interpretable visualizations and generating insightful outputs at each layer. The experiment utilized pedestrian perception frames from a traffic scenario as input images for two different VAE networks. The first network, \emph{VAE}${}_1$, was trained on a dataset of approximately 805k frames extracted from a fixed pedestrian route in the traffic scenario. The second network, \emph{VAE}${}_2$, was an enhanced version trained on a larger dataset of approximately 4m frames, extracted from different pedestrian routes covering a wider range of pedestrian perspectives. Our objective was to determine if our tool could justify the improved performance of \emph{VAE}${}_2$.
To evaluate the functionality of the two networks, we used the correlation distance metric $r(u,v) = 1 - \frac{(u - \Bar{u}) \cdot (v - \Bar{v})}{||u||_2~||v||_2}$ to group between filters $u$ and $v$ in the same layers similar filters of \emph{VAE}${}_1$ and \emph{VAE}${}_2$ and identify differences in extracted features between them. For each filter $u$ in a convolutional layer $L \in \{1,2,3,4\}$ in \emph{VAE}${}_2$, we paired it with the filter $v$ in the same convolutional layer in \emph{VAE}${}_1$ that minimized $r(u,v)$. This enables us to compare filters performing the same functionality across both VAEs.

\paragraph{Latent Space Interpretation}
\label{paragraph:latentSpaceMethod}

Besides our convolutional feature maps visualization tool that explains the internal functionality of the convolutional VAE, here,
we present a~complementary approach to explain the mapping between image features and latent vector values (Deep Network Processing~\cite{xaiconcepts}).
Inspired by~\cite{Ha:2018:RWM:3327144.3327171} and \cite{vae}, we observe how systematic changes to latent values influence the visual features of the decoded latent vectors. 
Such mapping between latent vector changes and decoded visual features will later help us to explain the LSTMs (see Section~\ref{paragraph:lstmMethod}).
To approach this, we design an experiment where we systematically manipulate the 50D latent vector values of $30$ encoded traffic scenario frames, extracted from the pedestrian scenario mentioned in Section~\ref{section:intro}.
Our initial experiments indicate that manipulating the values at one or a few positions of the latent vector results in minute changes at the decoded frames.
Therefore, we divide each latent vector into five equally sized regions of size ten. We arrived at this division through qualitatively experimenting several division setting and choosing the most interpretable one.  %
We test the influence of iteratively interpolating the values of a region by increments of $\{1,2,3\}$. After analyzing the influence of one region, we restore the values of the original vector, then we switch to the next region. This results in three different analyses per region and 15~frame analyses in total.
Finally, the manipulated latent vectors are decoded back to $45\times85$ pixel images that are stored in a~$3\times5$ grid that we call the \emph{latent grid}. We apply the latent grid visualization to \emph{VAE}${}_1$ introduced in Section~\ref{paragraph:convVaeMeth}.

\paragraph{RG-inspired Interpretable Autoencoder Architecture}

Most NNs are inherently non-transparent, meaning interpreting their functionality requires external analysis~\cite{xaiconcepts}.
So far, we have presented two external analysis tools that visualize the internal functionality of convolutional layers and examine the mapping between latent encoding and image features.
However, those techniques provide a limited interpretation of the mapping between the latent and feature space of the ConvVAE, a common issue in interpreting black-box models~\cite{stopExplainingBlackBox}.

Here, we present a different approach, where instead of explaining such a~non-transparent network, we reconstruct it into a model that is designed to be transparent and interpretable by nature that we can use as a backbone for an interpretable VAE architecture.
First, we apply the SVD on a sample of the training data to extract interpretable singular vectors~$U$ that represent the most relevant features.
Each singular vector corresponds to a principal component with the same dimensionality as the dataset frames. Consequently they can be visualized. 
We use the 2D Fast Fourier Transform (FFT) on the singular vectors~$U$, filter out higher frequency modes with a low pass filter, which is equivalent to applying an RG transformation~\cite{koch2} to remove irrelevant information, then apply the Inverse 2D Fourier transform to return to the singular vector space. The resulting filtered singular vectors matrix~$\alpha$ is used to encode an input image into a latent vector~$z$ by projecting it onto a lower dimensional vector space with the basis of singular vectors~\cite{svd}. The values of~$z$ indicate the degree of similarity between the input image and the singular vectors, where higher values indicate a stronger presence of features represented by the singular vectors in the input image.
Similarly, $z$~can be decoded back to an image using the transpose of the singular vector matrix $\alpha^T$ via 
$
    z = \alpha^T x,~
    \hat{x} = \alpha z.
$

The hyper-parameters of the pipeline are the number of singular vectors used in the matrix~$\alpha$ that ultimately determine the size of the latent vector~$z$--since the input image is projected onto every single vector--and the cutoff frequency of the low pass filter. 
Since each singular vector can be visualized as a $45 \times 85$ frame, the features that %
$z$ encodes can now be mapped to a~human understandable domain. We test the pipeline using a subset of the training dataset of \emph{VAE}${}_1$ consisting of 16,500 pedestrian perception frames. We use the Kullback-Leibler (KL) divergence~\cite{kullback1951information} to evaluate the pipeline's performance at a range of hyperparameter settings and to compare against the baseline VAE. We showcase the pipeline's interpretability by visualizing singular vectors of $\alpha$ demonstrating abstract features encoded by~$z$. %

\paragraph{Interpreting LSTM Dynamics and Feature Relevance}
\label{paragraph:lstmMethod}

The LSTM architecture's memory cell is a crucial component that stores state information from previous prediction steps. At each step, the memory cell decides whether to forget its current state and replace it with new input or maintain the current state~\cite{lstmoriginal}. Hidden state values~$h_c$ of the memory cells determine whether to affect the network's output at each step. The memory cell's decisions are regulated by gates, which are simple multiplicative units that pass through a \emph{tanh} or sigmoid activation function~\cite{lstmoriginal}. The LSTM's design enables it to store important information from prior states while protecting them from irrelevant inputs or noise~\cite{lstmoriginal}. %
To explain both the internal functionality and the feature relevance of the pedestrian perception prediction model from \cite{petzold2022if}, we present two approaches for calculating the relevance of input features and for interpreting the functionality of LSTM memory cells in a street-crossing scenario, respectively. In the crossing scenario, a pedestrian approaches a zebra crossing, uses it to cross the street, and then continues on their way. The scenario is executed in CARLA~\cite{Dosovitskiy17} and the pedestrian's semantically segmented vision and action commands are recorded. At each time step, the LSTM uses a pedestrian action $a_t$ and latent vector $z_t$ to predict the next perception frame $z_{t+1}$, which is passed back to the LSTM in a feedback loop to predict again. 
The first approach is based on an input of 400 frames of the pedestrian's vision and storing the hidden states of the LSTM's 512 memory cells at each frame prediction. We use a sigmoid function to normalize the hidden state values between zero and one, where values closer to one indicate that the cell is triggered, that is, it passes its contents to the output. To visualize the activity of a hidden state with respect to the predicted frame, we plot the hidden state on the y-axis and the index of the generated frame on the x-axis in a 2D plot. We aim to identify cells with interpretable behavior (i.e., correlating the cells' hidden values and the features of the pedestrian perception) using two heuristic filters.

We qualitatively evaluate the interpretability of the top cells identified by the two filters based on a human user's ability to correlate the cell's hidden state behavior with the events in the traffic scene and the actions of the ego-pedestrian. Our investigations indicate that cells with noisy hidden state behavior tend to be uninterpretable, while interpretable cells tend to be active only at certain intervals.

We define the first filter~$\kappa$ that returns cells that minimize the KL divergence between their hidden state values and a square pulse activated at a particular range when a certain event occurs, such as when the pedestrian passes the crosswalk:
\begin{equation}
    \kappa = \arg \min_c (\text{KL}(h_c || \theta(t-r_1) - \theta(t-r_2)))\;,
\end{equation} 
where \emph{KL}$(\cdot||\cdot)$ is the KL divergence, $r_1$~and~$r_2$ are the start and end points of the interval~$I$ under investigation, $c$ is the index of a memory cell, $h_c$~is the hidden state values of cell~$c$, and $\theta$ is the Heaviside step function.
The LSTM takes in the ego-pedestrian actions $a=[a_0,a_1,a_2]$, where $a_0$ indicates movement (0 for stopped, 1 for in motion), $a_1$ is the pedestrian's body angle parallel to the ground, and $a_2$ is the head's rotation parallel to the ground.
We introduce the second filter $\mu$ that identifies cells tracking action changes:
\begin{equation}
\mu = \arg \min_c (S_c(\nabla h_c, \nabla a))\;,
\end{equation}
where $S_c$ is the cosine similarity, $\nabla h_c$ is the gradient of hidden values $h_c$, and $\nabla a$ is the gradient of the action $a$. \\
In our second approach, we propose an augmented Layer-wise Relevance Propagation~\cite{lrp} technique to visualize the relevance of the input latent features and map them to the RGB space. After the LSTM model predicts the future latent vector, we assign a relevance value of $1$ for all values in $z_{t+1}$. 
Then we redistribute the relevance for lower layer neurons through two propagation rules. The two rules handle weighted connections and multiplicative connections such as in the cell state~(see~\cite{lrprules} for the formulas of our chosen rules).
Similar to the latent space interpretation (see Section~\ref{paragraph:latentSpaceMethod}) experiment, in order to map the relevance scores of the input latent space to the RGB space, we perturb each latent value and assign its relevance score to the most affected pixel regions in the decoded image. We use the image relevance scores as a mask to visualize the most relevant visual features of the input image. To evaluate the LRP explanation and the differences between the decision making process of the LSTM and other human or machine models, we compare the resultant masks to human driver attention maps generated by a pretrained model proposed by~\cite{driverattention} predicting the visual attention of human drivers. 
As metrics we use the mean NSS and Pearson's Coeffecient~\cite{driverattention} over the whole scenario. %
As input to the LRP experiment we use the pedestrian crossing scenario shown in~Table~\ref{table:frames} and the latent grids produced in Section~\ref{paragraph:latentSpaceMethod}. We test the LSTM's response to varying input features and unforeseen scenarios that may lead to traffic hazards.

\section{Results}

We present the results of our adaptation of the  convolutional feature map visualization~\cite{cnnviz} and latent space visualization~\cite{vae,Ha:2018:RWM:3327144.3327171} 
to implement both Deep Network Representation and Deep Network Processing~\cite{xaiconcepts}. We also show our novel explanation by simplification \cite{xaiconcepts} method to produce a transparent VAE, and our LSTM explainability framework combining memory cell analysis with latent space visualization. 

\begin{figure}[t]
    \centering
    \includegraphics[width=0.5\textwidth,trim={0 0 0 0.3cm},clip]{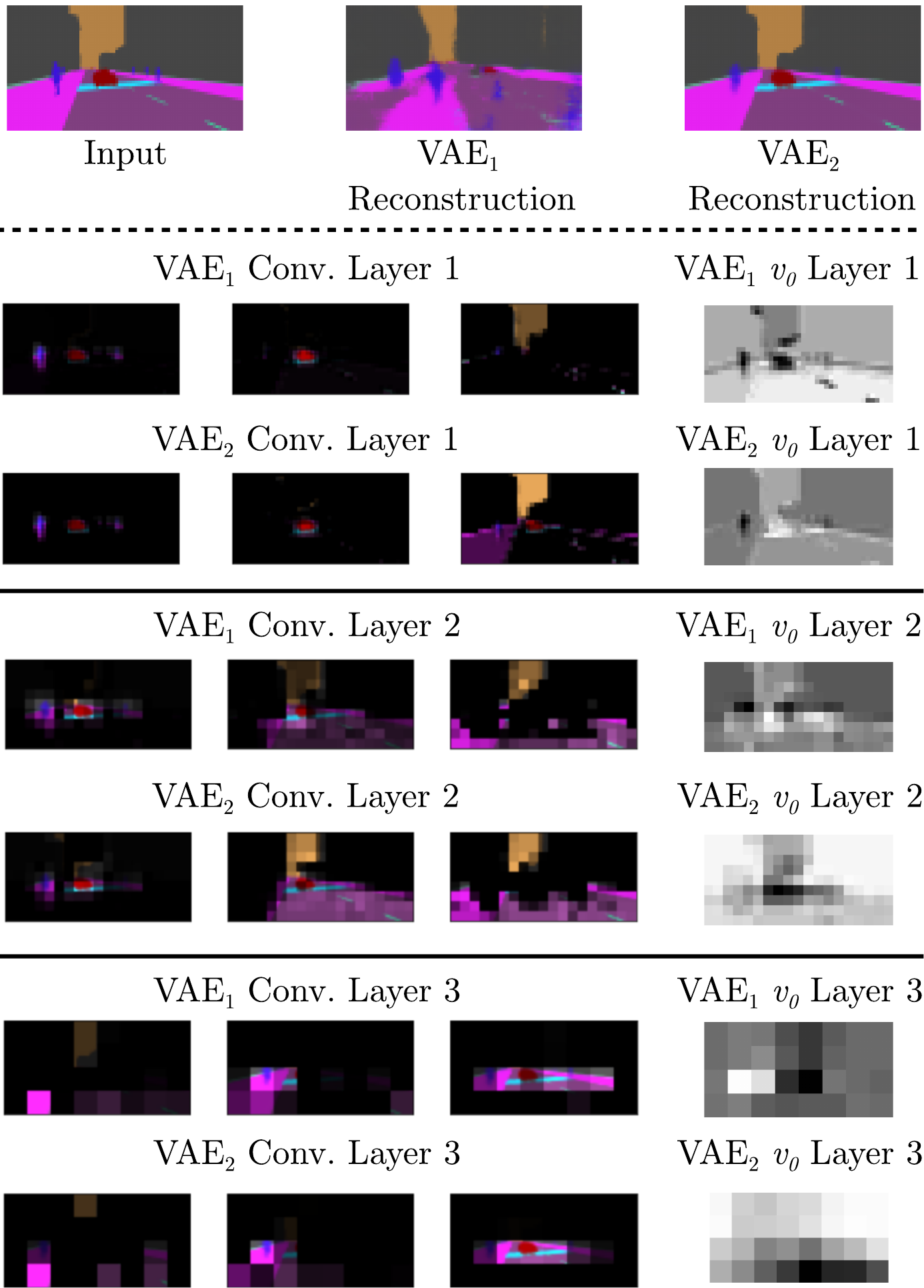}
    \caption{Visualization of the RGB feature maps and top singular vector $v_0$ for layers 1-3 for \emph{VAE}${}_1$ and \emph{VAE}${}_2$. For each layer, each column represents the VAEs' most similar feature maps.
    }
    \label{fig:convResults2}
\end{figure}

\paragraph{Convolutional VAE Feature Visualization}
We present the visualizations of the feature map comparison discussed in Section \ref{paragraph:convVaeMeth}. Figure~\ref{fig:convResults2} shows that unlike \emph{VAE}${}_2$, \emph{VAE}${}_1$ was unable to effectively reconstruct the input, since the decoded image is noisy with missing features, such as the car and the crosswalk. 
In the bottom of Figure~\ref{fig:convResults2}, each column represents the two pairs of feature maps belonging to filters in the same layer in the two VAEs.
In the first layer, \emph{VAE}$_2$ better captures the skyline feature as indicated by its brighter color.

In the visualization of the second layer and third convolutional layers, we observe similar trends, with \emph{VAE}${}_1$ poorly extracting the features of the crosswalk and skyline compared to \emph{VAE}${}_2$. These visualizations indicate that the filter of \emph{VAE}${}_1$ have not learned to properly extract the skyline and crosswalk. We exclude the $2\times2$ fourth layer's feature map comparison and singular vectors, as their low dimensionality lacks interpretability and hinders post-hoc interpretability in networks with chained convolutional layers.
The singular vectors provide a method for users to see the effect of internal data transformations at each convolutional layer, especially in earlier layers with relatively larger filter dimensions. The first layer singular vectors of \emph{VAE}${}_2$ feature high activation values (marked by brighter colors) in patches corresponding to the crosswalk and skyline. Overall, the two-phase feature visualization pipeline %
has qualitatively demonstrated its ability to visualize internal data processes of the ConvVAE, visualizing internal causes of poor reconstruction by the ConvVAE, which could lead to dangerous traffic situations.

\begin{figure}[ht]
  \centering
  \subfloat{\includegraphics[width=0.4\textwidth]{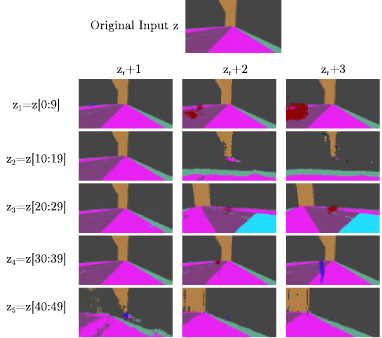}}
  \hfill
  \caption{Latent grids for $z$ (top). The rows represent regions of size ten in the latent vector. Each entry is the decoded vector taken after increasing the values of the region by $\{1,2,3\}$. }
  \label{fig:latentResults}
\end{figure}

\paragraph{Latent Space Interpretation}
\label{paragraph:latentSpaceResults}
We present here the $ 5 \times 3 $ latent grids of \emph{VAE}${}_1$ 
for one example of the 30 input latent vectors we manipulate, shown in Figure~\ref{fig:latentResults}. 
Each row in the grid, corresponding to a different latent vector slice of size ten, appears to control a set of semantic features of the decoded images. For example, the fourth row $z_4$ features the addition of a pedestrian, while the first row in features the addition of a car. By observing the different rows of the latent grid for a given image, a user can visualize a mapping between semantic features and their latent representation. The latent grid tool will be crucial in analyzing responses of the LSTM, since the latent vector is used by the LSTM to predict the next pedestrian perception frame.

\begin{figure}[htbp]
  \centering
  \subfloat{\includegraphics[width=0.3\textwidth]{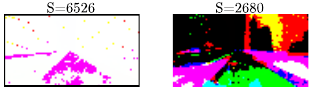}}
  \hfill
  \subfloat{\includegraphics[width=0.14\textwidth]{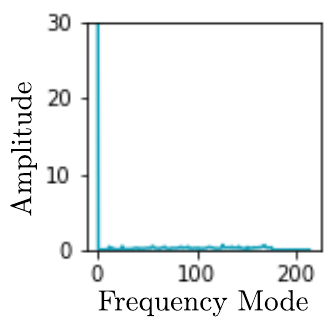}}
  \subfloat{\includegraphics[width=0.2\textwidth]{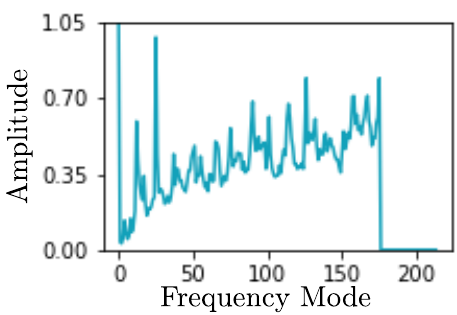}}
  \caption{Singular vectors of $\alpha$ with a cutoff frequency of 175 with the two highest singular values~$S$~visualized (top). We show the frequency domains of the first (bottom left) and twelfth singular vector~(bottom right). $\alpha[0]$ shows high support in only the lowest frequency, while $\alpha[11]$ has support throughout the spectrum, in agreement with~\cite{koch2}.
  }
  \label{fig:svd_results2}
\end{figure}

\paragraph{RG-inspired Interpretable Autoencoder Architecture}
For the SVD-based pipeline, we sample the data from the pedestrian vision dataset by~\cite{petzold2022if} to provide an input space of semantically segmented images. 
We test two cutoff frequencies set at 150 and 175 frequency modes. We compute the SVD using Tensorflow 2.9.1 on three Nvidia A100 GPUs. We rely on the implementation of~\cite{koch2} for the low pass filter, but refactor it using Tensorflow.
We compare the performance of SVD architectures to the baseline \emph{VAE}${}_1$ model using the KL Divergence reconstruction error of the test dataset of the \emph{VAE}${}_1$ model trained by~\cite{petzold2022if}. The \emph{VAE}${}_1$ model, the SVD autoencoder with cutoff of $175$ and $150$ have a mean reconstruction error of $0.024$, $0.179$, and $0.521$, respectively.
In terms of interpretability, we can visualize and interpret features encoded by the latent vector as each latent value represents the strength of a principal component~(the singular vector). For example, the top three singular vectors of the 175 frequency mode SVD in Figure~\ref{fig:svd_results2} show different perspectives of the ego-pedestrian. The first three values of the latent vector numerically correspond to each of these visualizable singular vectors, forming a
transparent model~\cite{stopExplainingBlackBox}.

\begin{figure}[htbp]
    \centering
    \begin{tikzpicture}[scale=0.5]
        \begin{axis}[
            xlabel={Frame Number},
            ylabel={Hidden State Value},
            cycle list name=color-list,
            tick pos=left,
            legend style={cells={anchor=east}, legend pos=north east, font=\large},
            xlabel style={font=\large},
            ylabel style={font=\large},
            trim axis left,
            trim axis right, 
            name = plot1]
            \addplot+ [line width=2pt, mark='-'] table [x=frame, y=isMoving Action, col sep=comma] {cell134.csv};
            \addplot+ [line width=2pt, mark = '-'] table [x=frame, y=hidden_cell_value, col sep=comma] {cell134.csv};
        \end{axis}
        
        \begin{axis}[
        cycle list name=color-list2, name=plot2, at={(plot1.south east)}, ytick=\empty, at={(plot1.south east)}, legend style={cells={anchor=east}, legend pos=north east, font=\large, xlabel={Frame Number} },
            xlabel style={font=\large},
            ylabel style={font=\large},
            trim axis left,
            trim axis right]
            \addplot+ [line width=2pt, mark='-'] table [x=frame, y=Head Angle Action, col sep=comma] {cell100.csv};
            \addplot+ [line width=2pt, mark='-'] table [x=frame, y=hidden_cell_value, col sep=comma] {cell100.csv};
        \end{axis} 
    \end{tikzpicture}
    \caption{ Cell 134 (Left) and Cell 100 (Right) show the hidden state on the y-axis in blue relative to the pedestrian vision frames. The pink signal is movement action $a_0$, while the cream signal is head angle action $a_2$. }
    \label{fig:lstmResult1}
\end{figure}
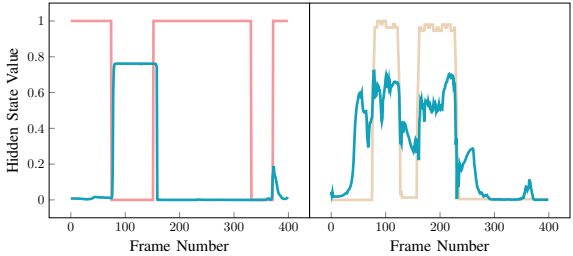

\paragraph{Interpreting LSTM Dynamics and Feature Relevance}
\label{paragraph:lstm1results}
We do two LSTM memory cell tests. First we analyze the LSTM memory cells as the VAE-LSTM model predicts the perception of an ego-pedestrian crossing the street. Table~\ref{table:frames} shows the three actions of the pedestrian and a textual description of the scene. We present a sample of memory cells with interpretable hidden state behavior, detected using filters~$\kappa$ and~$\mu$. Of the 512 cells, only 82 cells were found to meet the qualitative interpretation standard described in Section~\ref{paragraph:lstmMethod}. 
Cell 134 (Figure~\ref{fig:lstmResult1}-a) is active at range 80 to~159, corresponding to the interval where the ego-pedestrian stops walking while changing the angle of the head towards the street. Cell 134 was detected through the filter~$\kappa$ using a Heaviside square pulse between 80 and~159. The interpretation of this behavior is that cell 134 keeps track of the periods in the sequence when the pedestrian performs the `stop and scan' behavior before crossing the street, effectively acting as a movement detector. 
As for the filter~$\mu$ that detects cells sensitive to the three action values, no interpretable cells were found to react to differences in the first two action values, corresponding to the flag of the ego-pedestrian movement and the angle of the body of the ego-pedestrian.
However, cell 100 (Figure~\ref{fig:lstmResult1}-c) was found to react to changes in the third action value that represents the angle of head of the ego-pedestrian. Cell 100 has high hidden cell values coinciding with high normalized values of the head angle, when the pedestrians looks sideways toward the street and passing cars.
\begin{table*}[t]
\centering
\begin{tabular}{ccccc} 
 \toprule
{Frames} & {$a_0$} & {$a_1$} & {$a_2$} & {Scene description}\\ \midrule
0-80 & 1 & 180 & 0 & Walk(Sidewalk), LookTo(Sidewalk)\\
81-118 & 0 & [180 - 270] & [0 , -48.63] & Turn(Right), LookTo(Street)\\
119-135 & 0 & 270 & [-48.63,-0.66] & Turn(Head,Right), LookTo(Crosswalk)\\
136-158 & 0 & 270 & -0.66 & LookTo(Crosswalk)\\
159-233 & 1 & 270 & [-0.66 - 85.54] & Walk(Crosswalk), LookTo(Street)\\
234-337 & 1 & 270 & [85.54 - 0.24] & Walk(Crosswalk), LookTo(Crosswalk)\\
338-377 & 0 & [270-180] & 0.24 & Turn(Body,Left), LookTo(Sidewalk)\\
378-399 & 1 & 180 & 0.24 & Walk(Sidewalk), LookTo(Sidewalk)\\
\bottomrule
\end{tabular}
\caption{ \label{table:frames} A pedestrian scenario description matching time frame ranges with pedestrian perception. }
\end{table*}

\begin{figure}[ht]
    \centering
    \includegraphics[width=0.45\textwidth]{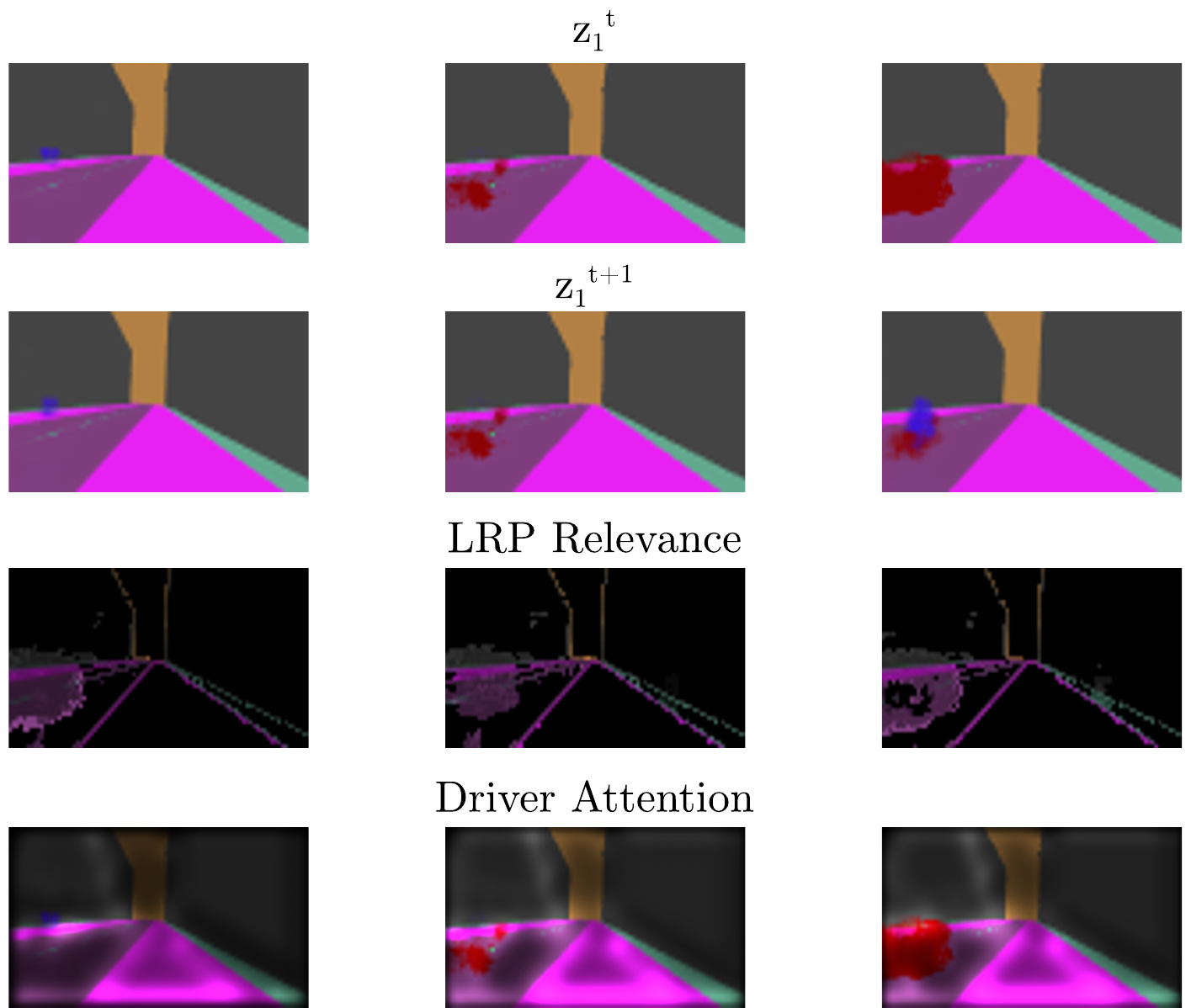}
    \caption{LSTM predictions in second for latent grid $z_1$, with the LRP heatmap and human attention map in the third and fourth rows, respectively. The LSTM generates a cyclist instead of car, with the LRP detecting the presence of the car's region.}
    \label{fig:lstmResult2}
\end{figure}

In our second LSTM approach, we visualize and evaluate the input relevance heat maps using LRP augmented with perturbation masking for the pedestrian scenario frames of~Table~\ref{table:frames} and the latent grids as the example shown in~Figure~\ref{fig:latentResults}. The empirical evaluation of the comparison between the LRP heatmaps and driver attention maps resulted in a mean NSS score of 0.53 and a Pearson Coeffecient of 0.46 for the pedestrian scenario of~Table~\ref{table:frames}. Using the latent grid as input to the prediction model, we were able to detect unpredictable behavior of the LSTM. In the next frame prediction of $z_{1}+2$ and $z_{1}+3$ in Figure~\ref{fig:lstmResult2}, the LSTM changes the car in the input frame to a cyclist in the output frame, which may cause dangerous effects in a DAS relying on future pedestrian perception. The LRP heatmaps for this scenario indicate that the latent manipulation of that region triggered the response of the LSTM even before the car becomes visible in frame~$z_{1}+1$.

\section{Discussion}

Our four methods apply several XAI techniques to ConvVAEs and LSTMs.
The use case we address is a vision model that gives drivers recommendations based on pedestrian perception. We can use our feature map visualization tool to investigate internal deficiencies that cause poor model performance.
The latent grid 
maps between latent space and feature space help to visualize the effect of changing different input features on the LSTM prediction in Section~\ref{paragraph:lstm1results}. A~drawback of our latent grid approach is its high granularity, as it does not interpret each latent vector value. 
Our prediction model explanation provides a baseline to examine the internal functionality and  safety of the LSTM predictions prior to deployment in a DAS. Both the memory cell and LRP analysis can be applied to other prediction architectures such as Transformers~\cite{transformers} by analyzing the attention head activity and designing propagation rules for the different Transformer layers. By creating scenarios with rare or dangerous features, we can analyze prediction models' dynamics and the relevance of critical features. However, a drawback of our evaluation is the use of the driver attention model trained to simulate drivers' rather than pedestrians' point of view. 
Finally, we proposed a backbone for an interpretable VAE architecture. Inspired by the RGM~\cite{koch2}, the ultimate goal is to fine-tune the SVD pipeline as an interpretable backbone of a VAE. The singular vectors of the reconstruction matrix~$\alpha$ can be used without post-hoc explanation to provide a mapping between feature and latent space. However, calculating the SVD for large datasets seems computationally infeasible. Our use of SVD to encode and decode images is a linear Principal Component Analysis (PCA), which is sensitive to outliers and corrupted data~\cite{svd}.

\section{Conclusion and Future Work}

We introduced an end-to-end XAI framework for explaining the internal functionality and feature-output mapping of DL models in a perception prediction network for autonomous driving~\cite{petzold2022if}. We uncovered the internal feature extraction process of the Convolutional VAE with convolutional feature map visualization and adopted a Feature Relevance explanation approach relating the feature space with the latent space through the latent grid analysis. Using the latent grids as inputs helped us to detect significant prediction errors by the LSTM that may affect the safety of VRUs, with our LRP explanation providing explanations of the relevant features impacting predictions. We have detected a significant space of interpretable LSTM memory cells and proposed methods to infer their functionality, such as keeping track of pedestrian movement, the presence of cars, and street crossing. Our empirical comparison of the LRP and driver attention maps presents an interpretation baseline for pedestrian prediction models and can be improved by training a pedestrian attention model to compare to the prediction model feature relevance. Our XAI methods may be beneficial in evaluating the transparency and trust-worthiness of state-of-the-art DL models in AD, enabling their certification in the future.
Our autoencoder architecture with an inherently interpretable latent space is based on the relationship between autoencoders, PCA, and RG~\cite{koch2}. For future work, we plan to integrate the SVD pipeline into a VAE and combine convolutional layers with the SVD to create hybrid VAE models with a basis of interpretability. We plan to use incremental and randomized SVD~\cite{svd} to increase dataset size and Robust PCA~\cite{svd} to enhance generalization and noise handling.

\bibliographystyle{IEEEtran}
\bibliography{bib}
\end{document}